\documentclass[10pt,twocolumn,letterpaper]{article}

\usepackage{cvpr}
\usepackage{times}
\usepackage{epsfig}
\usepackage{graphicx}
\usepackage{amsmath}
\usepackage{amssymb}
\usepackage{mdwlist,hhline} 

\usepackage[pagebackref=true,breaklinks=true,letterpaper=true,colorlinks,bookmarks=false]{hyperref}

\cvprfinalcopy 

\def\cvprPaperID{763} 

\begin{document}

\title{An Egocentric Look at Video Photographer Identity}

\author{Yedid Hoshen ~~~~~~~~~~~~~~~~~~~ Shmuel Peleg\\
The Hebrew University of Jerusalem\\
Jerusaem, Israel
}

\maketitle

\begin{abstract}
Egocentric cameras are being worn by an increasing number of users, among them many security forces worldwide. GoPro cameras already penetrated the mass market, reporting substantial increase in sales every year. As head-worn cameras do not capture the photographer, it may seem that the anonymity of the photographer is preserved even when the video is publicly distributed. 

We show that camera motion, as can be computed from the egocentric video, provides unique identity information. The photographer can be reliably recognized from a few seconds of video captured when walking. The proposed method achieves more than 90\% recognition accuracy in cases where the random success rate is only 3\%.

Applications can include theft prevention by locking the camera when not worn by its rightful owner. Searching video sharing services (e.g. YouTube) for egocentric videos shot by a specific photographer may also become possible. An important message in this paper is that photographers should be aware that sharing egocentric video will compromise their anonymity, even when their face is not visible.
\end{abstract}

\section{Introduction}
\label{sec:intro}

The popularity of head worn egocentric cameras is increasing. GoPro reports an increase in sales of 66\% every year, and cameras are widely used by extreme sports enthusiasts and by law enforcement and military personnel.

Special features of egocentric video include:
\begin{itemize*}
\item The camera is worn by the photographer, and is recording while the photographer performs normal activities.
\item The camera moves with the photographer's head. 
\item The camera does not record images of the photographer. In spite of this we show that photographers can often be identified.
\end{itemize*}

\begin{figure}[tb]
\centering{
\includegraphics[height=0.15\textwidth]{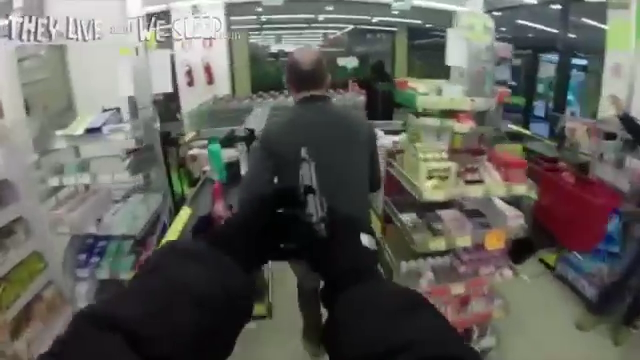}
\includegraphics[height=0.15\textwidth]{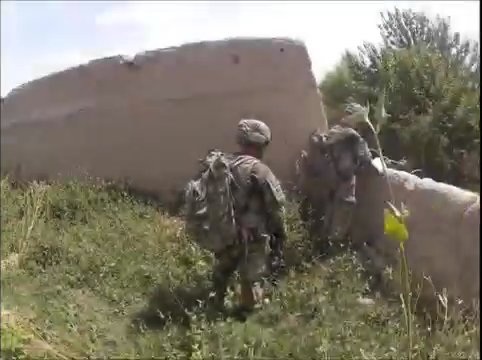}\\
a)~~~~~~~~~~~~~~~~~~~~~~~~~~~~~~~~~~~~~~~~~~~~~~
b)
\caption{a) A GoPro video uploaded to YouTube allegedly capturing a crime from the POV of the robber. Can the robber be recognized?  b) A GoPro video uploaded by US soldiers in combat. Are their identities safe? 
}
\label{Fig:PR}}
\end{figure}

Photographers feel secure that sharing their egocentric videos on social media does not compromise their identity (Fig.~\ref{Fig:PR}). Police forces routinely release footage of officer activity and operations of special forces recorded by wearable cameras are widely published on YouTube. Some have even recorded and published their own crimes. A consequence of our work is that the photographer identity of such videos can sometimes be found from camera motion.

Body motion is an accurate and replicable feature for identifying people over time. It is often recorded by accelerometers (\cite{mantyjarvi2005identifying}) or by an overlooking camera. Egocentric video can effectively serve as a head mounted visual gyroscope and can accurately capture body motion information. It follows that any egocentric video which includes walking contains body motion information that can accurately reveal the photographer.

\begin{figure}[tb]
\centering{
\includegraphics[width=0.49\textwidth]{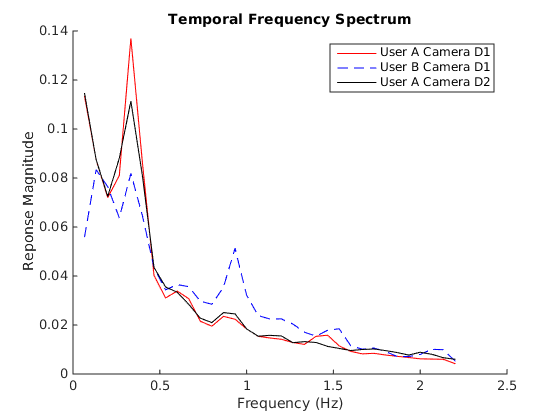}

\caption{Comparison of the temporal frequency spectra for three videos. Two videos were recorded using camera D1 by users A and B, the third video was recorded by user A using camera D2. It is readily seen that the spectra of the two videos recorded by photographer A are very similar to each other despite being recorded by different cameras and at different times. This suggests that a photographer's physique is expressed in the motion observed in his video.}
\label{Fig:Spectrum}}
\end{figure}

Specifically, we use sparse optical flow vectors (50 flow vectors per frame) taken over a few steps (4 seconds). This results in a set of time-series, one for each component of each optical flow vector. In Fig~\ref{Fig:Spectrum} we show the temporal Fourier Transform of one flow vector for three different sequences, showing visible differences between different photographers.

As a first approach for determining photographer identity, we computed LPC (Linear Predictive Coding \footnote{The LPC coefficients of a time series are $k$ values that when scalar multiplied with the last $k$ measurements of the time series, will optimally predict the next measurement.}) \cite{campbell1997speaker} coefficients for each of the optical flow time series. All LPC coefficients of all optical flow sequences were used as a descriptor. Photographer recognition using a non-linear SVM trained on the LPC descriptor gave 81\% identification accuracy (vs. accuracy of 3\% in random) and verification EER (Equal Error Rate) of 10\%.

Our second approach learns the descriptor and classifiers using a Convolutional Neural Network (CNN) which includes layers corresponding to body motion descriptor extraction and to recognition. The CNN is trained on the optical-flow features described above. Using CNN improves the results over the LPC coefficients, yielding 90\% identification rate (vs. accuracy of 3\% in random) and verification EER (Equal Error Rate) of 8\%.


The above experiments were performed on both a small (6 person) public dataset \cite{fathi2012social} (originally collected for Egocentric Activity Analysis) and on a new, larger (32 person) dataset collected by us especially for Egocentric Video Photographer Recognition (EVPR). 

The ability to recognize the photographer quickly and accurately can be important for camera theft prevention and for forensic analysis (e.g. who committed the crime). Other applications are web search by egocentric video photographer and organization of video collections. Wearing a mask does not reduce recognition rate, of course.

\section{Previous Work}
\label{sec:prev}

Determination of the painter of an artwork for preventing forgery and fake artists has attracted attention for centuries. Computer vision researchers have presented several approaches for automatic artist and style classification mainly utilizing low-level and object cues \cite{johnson2008image, bar2014classification}. 

Recognizing the unseen photographer of a picture is an interesting related problem. In this setting the photographic style \cite{thomas2015s} and the location of the photograph \cite{hays2008im2gps,lee2015predicting} can be used as cues for photographer recognition. Both methods are unable to distinguish between photographers using cameras on default settings (such as most wearable cameras) and at the same locations. Another approach is automatic recognition of the photographer's reflection (e.g. in the subject's eyes \cite{nishino2006corneal}), but this relies on having reflective surfaces in the pictures.

Photographer recognition from wearable camera video is a novel problem. Such video is jittery due to the motion of the photographer's head and body. Although typically a nuisance, we show that frame jitters can accurately determine photographer identity. 

Human body motion was already used for recognition. Gait recognition is typically done by a video camera observing a person's shape and dynamic walking style. These features are able to recognize a person accurately \cite{murase1996moving}. In our scenario, however, the photographer is not seen by the camera which is worn on his head. Recognition from accelerometers carried on the user's body \cite{mantyjarvi2005identifying} is also reported. Shiraga et al. \cite{shiraga2012gait} studied people recognition wearing a backpack with stereo cameras. Rotation and period of motion were computed using 3D geometry, and users were accurately recognized. Unlike all prior art, we are interested in recognizing photographers of videos taken by standard wearable cameras (e.g. as exist on video sharing websites), nearly all of which are monocular, head or chest mounted.

Using optical flow for activity recognition from head-mounted cameras has been done by \cite{kitani2011fast,polegtemporal,ryoo2013first,lu2013story} and others. Papers \cite{poleghead,yonetani2015egosurf} used head motion to retrieve head-mounted camera users observed in other videos recorded at the \textit{same time}. We, on the other hand, use camera motion to recognize the users of wearable cameras \textit{across time}.

Feature design for time series data has been extensively studied, particularly for speech recognition systems (\cite{reynolds1995robust}). Speaker verification is a long standing problem which is related to this work. Linear Predictive Coding (LPC) descriptors are very popular for speaker recognition \cite{furui1981cepstral}. We show that an LPC-based descriptor is highly effective also for user recognition from egocentric camera video.

In this paper we also take an end-to-end approach of learning features along with the classifier, instead of hand designing the features. We perform this using convolutional neural networks (CNN). For an overview of deep networks see \cite{bengio2009learning}. Learned features are sometimes better than hand-designed features \cite{krizhevsky2012imagenet}.

\section{Photographer Recognition from Optical Flow}
\label{sec:method}
 
Egocentric video suffers from bouncy and unsteady motion caused by photographer head and body motion. Although usually a nuisance, we show that this motion forms the basis for accurate photographer recognition methods.  

We present our basic features in Sec.~\ref{subsec:pre}. Two alternative descriptors and classifiers are described in Sec.~\ref{subsec:lpc} and Sec.~\ref{subsec:cnn}. 



\subsection{Feature Extraction}
\label{subsec:pre}

\begin{figure}[tb]
\centering{
\includegraphics[width=0.48\textwidth]{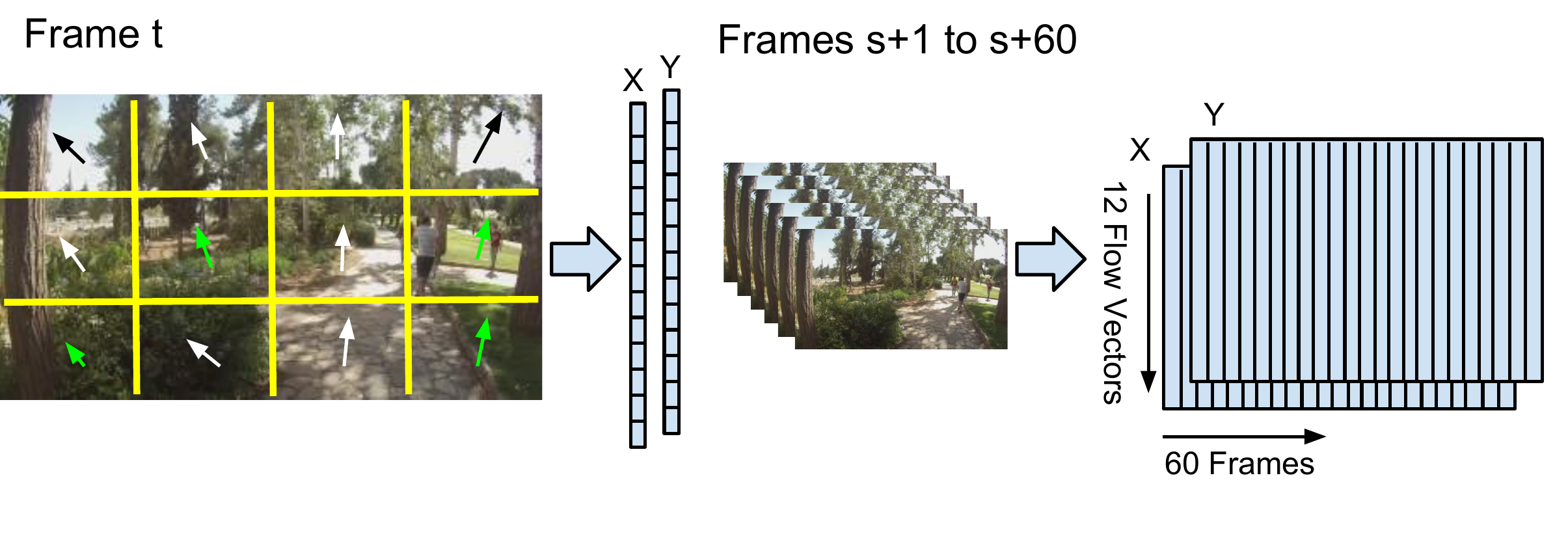}
a)~~~~~~~~~~~~~~~~~~~~~~~~~~~~~~~~~~~~~~~~~~~~~~~b)
\caption{a) 50 Optical flow vectors are calculated for each frame (only 12 shown here), and represented as two columns (each of 50 values), for the $x$ and $y$ optical flow components. b) The feature vector consists of optical flow columns for 60 frames, stacked into two 50$\times$60 arrays, for the $x$ and $y$ components of the flow.}
\label{Fig:Extraction}}
\end{figure}

\begin{figure}[tb]
\centering{
{\small Horizontal flow component ~~~~~~~Vertical flow component}\\
\rotatebox{90}{~~ $\longrightarrow$ ~~~ \rotatebox{270}{\it x}} 
\includegraphics[width=0.21\textwidth]{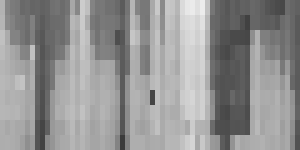} ~~
\includegraphics[width=0.21\textwidth]{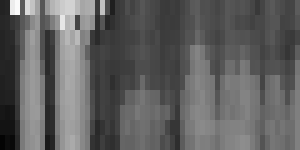}\\
$\longrightarrow ~~ t$ ~~~~~~~~~~~~~~~~~~~~~~~~~~~~~~~~~ $\longrightarrow ~~ t$ ~~~~~~~~~~~~~~~~~~~~~ \\
\rotatebox{90}{~~ $\longrightarrow$ ~~~ \rotatebox{270}{\it x}} 
\includegraphics[width=0.21\textwidth]{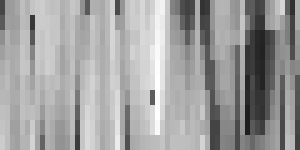} ~~
\includegraphics[width=0.21\textwidth]{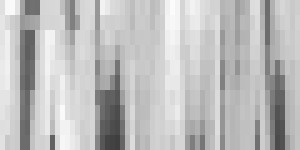}\\
$\longrightarrow ~~ t$ ~~~~~~~~~~~~~~~~~~~~~~~~~~~~~~~~~ $\longrightarrow ~~ t$ ~~~~~~~~~~~~~~~~~~~~~ \\

\caption{Two examples of the flow feature vectors. Each feature vector consists of 50 optical flow vectors per frame, computed for each of 60 frames. Here only the central row, having 10 flow vectors, is shown. The left and right images show the horizontal and vertical components of the optical flow. Note the rich temporal structure along the time axis.}
\label{Fig:FrontEnd}}
\end{figure}

In the following sections we assume that the video frames were pre-processed in the following way (see Fig.~\ref{Fig:Extraction}): 
\begin{enumerate*}
\item Frames are partitioned into a small number ($m_x \times m_y$) of non-overlapping blocks. 
\item $m_x \times m_y$ optical flow vectors are computed for each frame using the Lucas Kanade algorithm \cite{lucas1981iterative}. We use 10$\times$5 optical flow vectors per frame.
\item A block of $T$ seconds of such optical flow vectors is taken. We used $T=4$ seconds, which is long enough to include a few steps. At 15 fps this results in 60 frames.
\item Each feature vector covers a period of 4 seconds, and we computed feature vectors every 2 seconds. There is an overlap of 2 seconds between two successive feature vectors.
\end{enumerate*}

We used optical flow features for photographer recognition, rather than pixel intensities, as the body motion is eventually expressed by the pixel motion. On the other hand, recognition should be invariant to the specific objects seen in the environment, objects that are represented by pixel intensities. CNNs may be able to learn optical flow from pixel intensities, but learning this will require much more data than we can collect.

If dense optical flow were used as a feature, the high feature dimensionality would have lead to overfitting on small datasets. Using a smaller number of flow vectors gave reduced accuracy. In looking for the optimal feature size we found out that a grid size of 10$\times$5  optical flow vectors was a good compromise between overfitting and accuracy.

The feature extraction process is shown in Fig~\ref{Fig:Extraction}. 
Visualization of two extracted feature vectors is shown in Fig.~\ref{Fig:FrontEnd}. Full details are in Sec.~\ref{sec:imp}.

\begin{figure*}[tb]
\centering{
\includegraphics[width=0.85\textwidth]{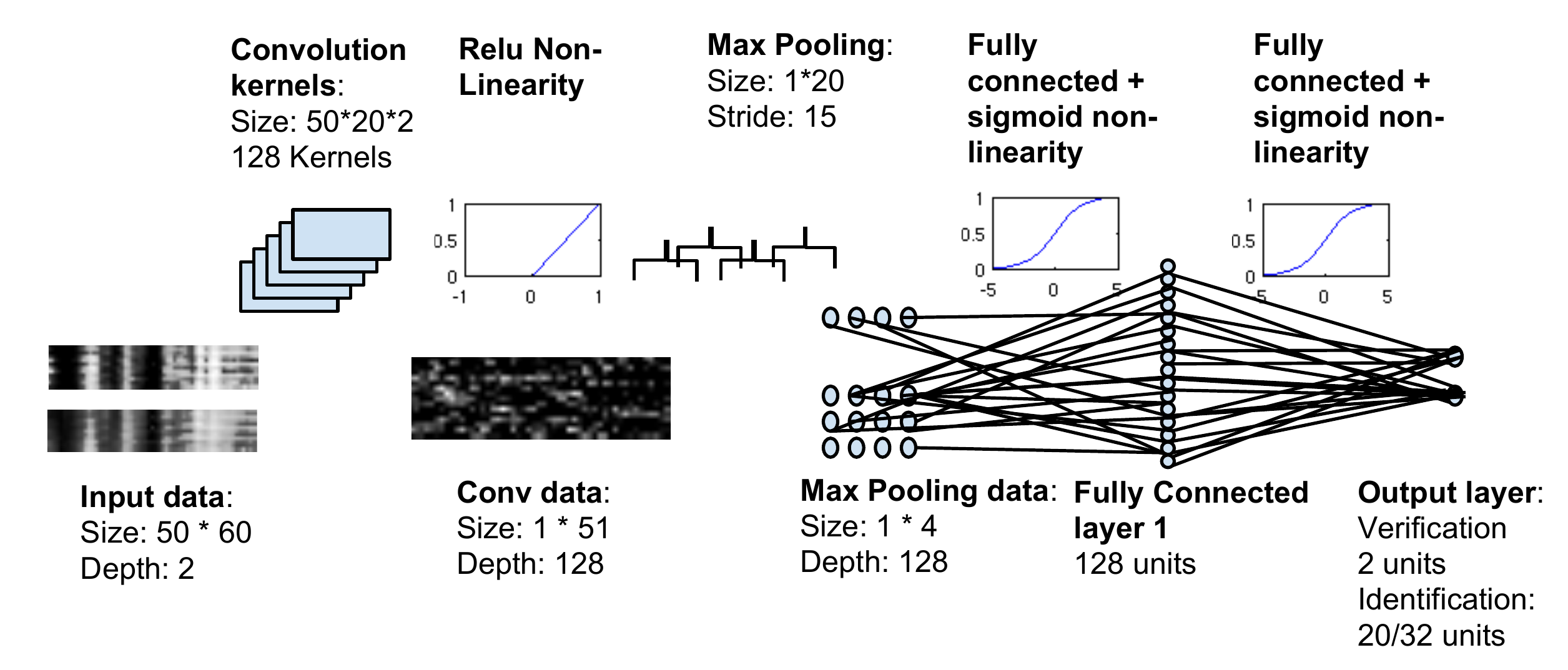}

\caption{A diagram of our CNN architecture for photographer recognition from a given flow feature vector. The operations on the data are shown on top, the sizes of subsequent data layers are shown on the bottom. The Neural Network learns the descriptor jointly with the classifier, therefore automatically creating a descriptor optimal to this task.}
\label{Fig:net_arch}}
\end{figure*}

\subsection{LPC Descriptor + Kernel SVM}
\label{subsec:lpc}

LPC \cite{campbell1997speaker} is a popular time-series descriptor (e.g. for speaker verification). LPC assumes the data is generated by a physical system, here the photographer's head and body. It attempts to learn a linear regression model for its equations of motion, predicting for each optical flow series the flow value in the next frame given the flow values of previous $k$ frames. Given a feature vector, we calculate an LPC model for each component of each 4s flow time series (100 models in total). Using too few coefficients yields less accurate predictions, while too many coefficients causes overfitting. We found $k$=9 to work well for our case. The final LPC descriptor consisted of all coefficients of all time-series models (100$\times$9).

An RBF-SVM classifier was used for learning both identification (classify LPC descriptor into 1 of $M$ known photographers) and verification (classify LPC descriptor into target photographer or  rest-of-the-world). The non-linear (RBF) classifier outperformed linear SVM in almost all cases. As mentioned before, photographer recognition using a non-linear SVM trained on the LPC descriptor gave 81\% identification rate (vs. random 3\%), and the verification EER (Equal Error Rate) was 10\%. 

\subsection{Convolutional Neural Network}
\label{subsec:cnn}

In Sec.~\ref{subsec:lpc} we described a hand-designed descriptor for identity recognition. The LPC descriptor suffers from several drawbacks:
\begin{itemize*}
\item The LPC regression model is learned for each time-series separately and ignores the dependence between optical flow vectors.
\item The LPC descriptor and SVM classifier are learned independently, the labels cannot directly influence the design of the descriptor. 
\end{itemize*}

To overcome the above drawbacks, we propose to learn a CNN model for photographer recognition. The CNN learns descriptor and classifier end to end, and is able to take advantage both of dataset labels and the dependence between features when calculating filter coefficients. The CNN is a more general architecture, the LPC descriptor is a subset of descriptors learnable by the network. 

Due to the limited number of data points available in our datasets, we limit our CNN to only 2 hidden layers. Using more layers increases model capacity but also increases over-fitting, and this architecture yielded the best performance. The architecture is illustrated in Fig.~\ref{Fig:net_arch}.

Our architecture is tailored especially for egocentric video. As we use sparse optical flow we do not assume much spatial invariance in the features (differently from most image recognition tasks). On the other hand the precise temporal offset of the photographer's actions is usually not important, e.g. the precise time of the beginning of a photographer's step is less important than the time between strides. Our architecture should therefore be temporally invariant. The first layer was thus designed to be convolutional in time but not in space. 

The kernel size spans all the blocks across the $x$ and $y$ components over $K_T$ frames (we use $K_T= 20$ which is a little longer than the typical step duration). The convolutional layer consists of $M$ kernels (we use $M=128$). The outputs of the kernels $z^1_m=W_m \ast x$ are passed through a ReLU non-linearity ($max(z^1_m,0)$). We pool the outputs substantially in time, as the feature vector is of high dimension compared to the amount of training data available. To correspond to the typical time interval between steps we use kernel length of 20 and stride of 15.

The data is then passed through two fully connected (affine) layers each followed by a sigmoid non-linearity ($\sigma(z)=\frac{1}{1+e^{-z}}$). The first fully connected hidden layer has $N_1$ hidden nodes (we used $N_1=128$). The output of this layer is the learned CNN descriptor.

The second fully connected layer is a linear soft-max classifier and has the same number of nodes as the number of output classes: 2 classes for verification, and 20 or 32 classes for identification.

\subsection{Joint Prediction from Several Descriptors }
\label{subsec:joint}

Sec.~\ref{subsec:lpc} and Sec.~\ref{subsec:cnn} described a method to train a photographer classifier on a short (4 seconds) video sequence. The video used for recognition is usually significantly longer than 4 seconds.

\begin{figure}[tb]
\centering{
\includegraphics[width=0.156\textwidth]{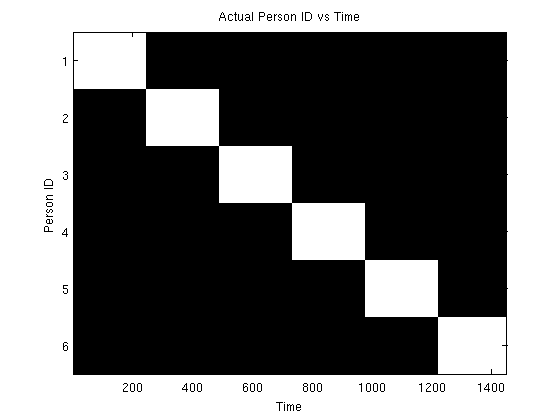}
\includegraphics[width=0.156\textwidth]{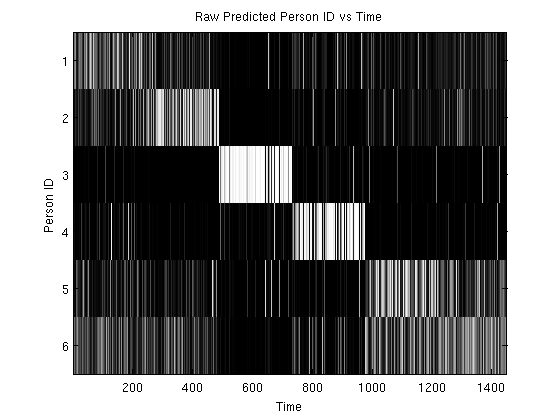}
\includegraphics[width=0.156\textwidth]{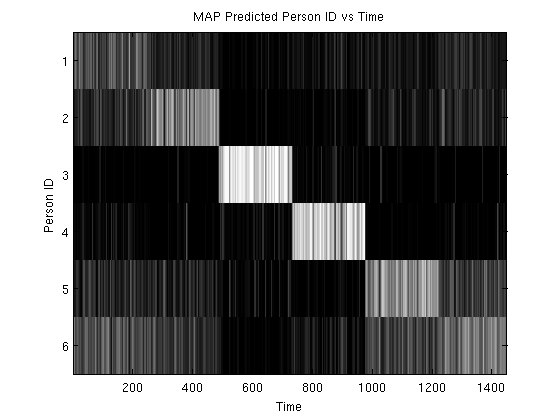}\\
a)~~~~~~~~~~~~~~~~~~~~~~~~~~~b)~~~~~~~~~~~~~~~~~~~~~~~~~~~~c)
\caption{The MAP rule operated on the FPIS dataset: a) Ground truth labels. b) Raw CNN probabilities. c) MAP rule probabilities (for $T$=12 seconds.). The MAP classifier visibly 'cleaned up' the prediction. }
\label{Fig:ClassRule}}
\end{figure}

We split the video into 4 second subsequences (overlapping by 2 seconds) and extract their feature vectors $V_t$ ($t$ is the subsequence number). We compute the identity label ($L_t$) probability distribution for each feature vector $V_t$ using LPC or CNN classifiers trained as described above, We then classify the entire video into the globally most likely label, $argmax_i{\prod_t{P(L_t=i| V_t)}}=argmax_i{\sum_t{log(P(L_t=i| V_t))}}$. 
While this classifier assumes that feature vectors are IID, we have found that this requirement is not necessary for the success of the method. See Fig.~\ref{Fig:ClassRule} for an example on the FPIS dataset. MAP classification has helped boost the recognition performance on the EVPR dataset to around 90\% (an increase of 13\%) over the 4s rate.

\section{Results}
\label{sec:results}

Several experiments were performed to evaluate the effectiveness of our method. As there is no standard dataset for Egocentric Video Photographer Recognition, we use both a small (6 person) public dataset - FPSI \cite{fathi2012social} that was originally collected for egocentric activity analysis. For each photographer - morning sequences were used for training, and afternoon sequences for testing.

In order to evaluate our method under more principled settings, we collected a new larger (32 person) dataset - EVPR - specifically designed for egocentric photographer video recognition. In the EVPR dataset all photographers recorded two 7 minute sequences (from which we extracted around 200 four second sequences each) on the same day with different head-mounted cameras (D1,D2) for training and testing. 20 of the photographers also recorded another 7 minute sequence with yet another camera (D3) a week later. Both datasets are described in detail in Sec.~\ref{sec:dsdes}. The detailed experimental protocol is described in Sec.~\ref{sec:proto}. 

\begin{figure}[tb]
\centering{
\includegraphics[width=0.30\textwidth]{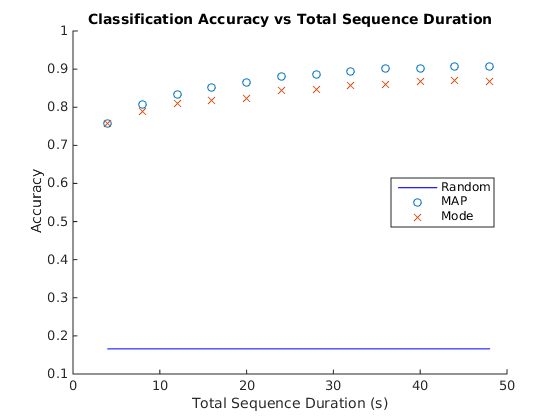}

\caption{Classification accuracy vs. video length when one feature vector covers $T=4$ seconds (Using CNN on the FPSI Dataset). Longer video allows extraction of more feature vectors. MAP classification consistently beats mode classification. Both methods can exploit longer sequences and thus improve on 4s sequence recognition. All methods perform far better than random.}
\label{Fig:fpis_class}}
\end{figure}

\subsection{Photographer Identification}
\label{subsec:iden_res}

Fig.~\ref{Fig:fpis_class} presents the photographer recognition test performance of our network on the FPSI database (6 people). The average correct recognition rate on a single feature vector (describing only 4 seconds of video) is 76\% against the random performance of 16.6\%. 

Test videos are usually longer than 4 seconds, and we have multiple feature vectors for each person. We combine predictions over a longer video using the MAP rule in Sec.~\ref{subsec:joint}. In Fig.~\ref{Fig:fpis_class} we compare the MAP strategy vs. taking the most frequent 4s prediction in the test video (Mode). We observe that using longer sequences further improves recognition performance, reaching around 91\% accuracy for 50 seconds of video. We also observe that MAP classifiers consistently beats the Mode classifier and use it in all other experiments.

\begin{figure}[tb]
\centering{
\includegraphics[width=0.30\textwidth]{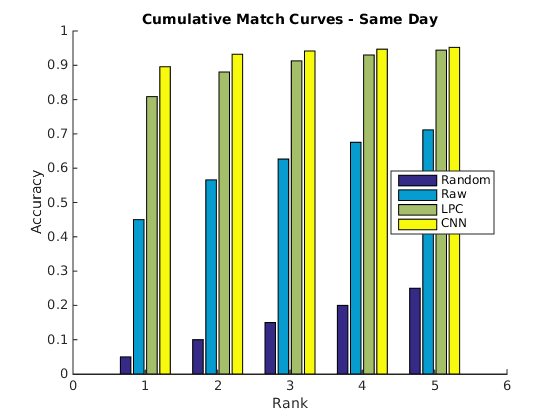}

\caption{CMC rates for same day recognition (for 12s sequences). LPC accuracy: 81\% (Top-1) and 88\% (Top-2). The CNN further improves the performance with 90\% (Top-1) and 93\% (Top-2). Both methods far outperform the random rate of 3\% (Top-1) and 6\% (Top-2). Both descriptors also beat the raw features by a large margin. }
\label{Fig:D2CMC}}
\end{figure}

\begin{figure}[tb]
\centering{
\includegraphics[width=0.30\textwidth]{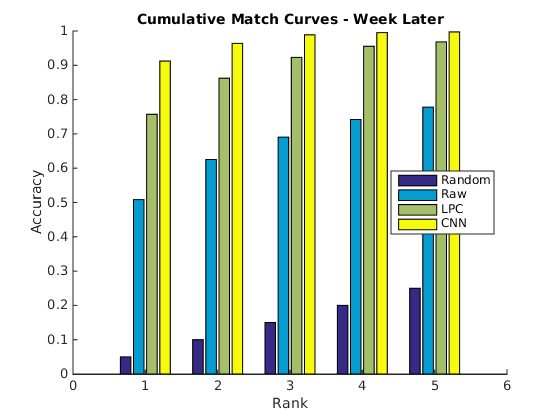}

\caption{CMC rates for recognition 1 week later (for 12s sequences). LPC accuracy: 76\% (Top-1) and 86\% (Top-2). The CNN further improves the performance with 91\% (Top-1) and 96\% (Top-2). Both methods far outperform the random rate of 5\% (Top-1) and 10\% (Top-2). Both descriptors also beat the raw features by a large margin. }
\label{Fig:WCMC}}
\end{figure}

To evaluate the recognition performance on a larger dataset, we show the performance of our method on our new dataset - EVPR. In this experiment the network was trained on video sequences for each photographer using Camera D1 and is evaluated on video sequences recorded on the same day using Camera D2 and a week later recorded using Camera D3. In Fig.~\ref{Fig:D2CMC} and Fig.~\ref{Fig:WCMC} we present the cumulative match curve (CMC) for the same day and week later recognition results respectively. We use the Top-$k$ notion, indicating that the correct result appeared within the top $k$ predictions of the classifier. In addition to LPC and CNN, an RBF-SVM trained on the raw optical flow features is used as baseline to evaluate the quality of our descriptors. High accuracy was achieved in both scenarios, same day CNN recognition accuracy is 90\% (top 1) and 93\% (top 2). The recognition performance a week later is better with 91\% (top 1) and 96\% (top 2). The improved performance numbers a week later are expected due to the smaller dataset size (20 vs 32), but are nonetheless encouraging as many photographers wore different shoes from the D1 training sequence recorded a week before. This result shows that our method can obtain good recognition performance on meaningful numbers of photographers and across at least a week.

\begin{table}[t]
\centering
\begin{tabular}{|l||c|c|c|c|}
\hline
 & \multicolumn{2}{|c|}{No Stab} & \multicolumn{2}{|c|}{Stab} \\
 \hhline{|~|--|--|}
Descriptor & 4s & 12s & 4s & 12s \\ \hline
LPC & 65\% & 81\% & 59\% & 72\% \\  
CNN & 77\% & 90\% & 71\% & 86\% \\ \hline
\end{tabular}
\caption{Same-day CSMC recognition accuracy with and without stabilization.}
\label{tab:stab}
\end{table}

To test the possibility that stabilization would take away some or all the body motion information in the frame motions, the identification experiments were redone with the following pre-processing stage: for each frame (50 flow vectors) the mean framewise vector was calculated and then subtracted from each of the vectors in the frame. As motion between frames is small and some lens distortion correction was performed, this is similar to 2D stabilization. Table.~\ref{tab:stab} shows that such "stabilization" degrades performance somewhat (4-9\%), but accuracy still remains fairly high. We note however that more complex stabilization might remove more body motion information. 

\subsection{Photographer Verification}
\label{subsec:ver_res}

We also test the verification performance obtained by our method. In order to evaluate verification performance by a single number it is common to use the Equal Error Rate (EER), the error rate at which the False Acceptance Rate (FAR) and False Rejection Rate are equal.

\begin{figure}[tb]
\centering{
\includegraphics[width=0.35\textwidth]{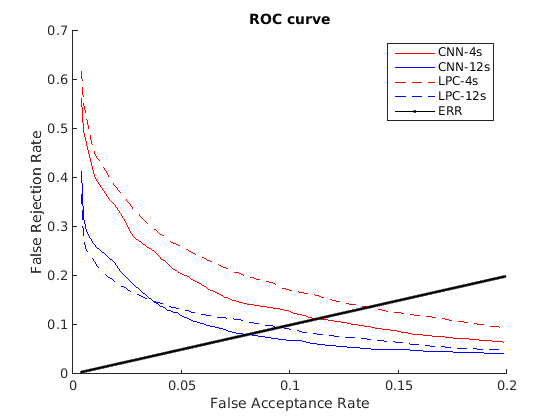}
\caption{ROC curves for the verification performance of our method for LPC and CNN descriptors of 4s and 12s sequences. For both methods we show the mean ROC curve. The EER of each method is given by the point of intersection between the linear line and its ROC curve. }
\label{Fig:VerROC}}
\end{figure}

The EER for both the CNN and LPC descriptors for videos of length 4s (one feature vector) and 12s (five feature vectors) is presented in Table.~\ref{tab:ver} while the ROC curves are shown in Fig.~\ref{Fig:VerROC}. A detailed description of our protocol can be found in Sec.~\ref{sec:pro}. It can be seen from our results that high accuracy (low EER) can be obtained by both descriptors: LPC 14\% (4s), 10\% (12s) and CNN 11\% (4s), 8\% (12s). The CNN obtains better performance for both durations with a larger improvement for 4s.

It should be noted that all test probe photographers apart from the target photographer had never been used in training. By focusing on modeling the target photographer we can separate him from the rest of the world, and are thus able to generalize to unseen test photographers. 

\begin{table}[t]
\centering
\begin{tabular}{|l||c|c|}
\hline
Descriptor & 4s & 12s \\ \hline
LPC & 13.6\% & 9.6\% \\  
CNN & 11.3\% & 8.1\% \\ \hline
\end{tabular}
\caption{Verification equal error rates for LPC and CNN descriptors with 4s and 12s sequence duration.}
\label{tab:ver}
\end{table}

\section{Discussion}
\label{sec:dis}

\begin{figure}[tb]
\centering{
{\small Horizontal flow component ~~~~~~~Vertical flow component}\\
\rotatebox{90}{ $\longrightarrow$ ~~ \rotatebox{270}{\it x}} 
\includegraphics[width=0.20\textwidth]{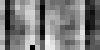}
\includegraphics[width=0.20\textwidth]{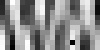}\\
$\longrightarrow ~~ t$ ~~~~~~~~~~~~~~~~~~~~~~~~~~~~~~~~~ $\longrightarrow ~~ t$ ~~~~~~~~~~~~~~~~~~~~~ \\
\caption{Examples of a temporal filter for the horizontal (left) and vertical (right) flow components. Horizontal axis is time, and vertical axis is location along the central line. The horizontal component filter appears to be sensitive for certain left-right frequencies, while the vertical component filter is sensitive to oscillating rotations: When the right side is moving up the left side is moving down, etc.}
\label{Fig:ShowFeatures}}
\end{figure}

\textit{Analysis of CNN features:} In order to analyze the features learned by the CNN we visualize the filters learned by the first layer. Fig.~\ref{Fig:ShowFeatures} shows the horizontal and veritical components of a first layer temporal filter learned by the network. For illustration purposes, only the weights of the central line of pixels are shown. Looking at the weights, we see that the horizontal component filter is tuned to respond to some specific frequencies, while the vertical component looks for sharp rotations. This behavior appears in several other filters suggesting that the network might be using both spectral and transitive cues.

\textit{Transfer Learning for verification:} In some scenarios it may not be possible to train a verification classifier for each photographer. In such cases Nearest Neighbors may be a good alternative. The following approach is taken: An identification CNN is trained on half the photographers in the training dataset. We choose a video by a target photographer (that was not used for training the CNN), and extract its CNN descriptors (as in Sec.~\ref{subsec:cnn}), this set of descriptors forms our gallery. Similarly we extract CNN descriptors from all video sequences of photographers not used for training the CNN, this forms our probe set (excluding the sequence used as gallery). For each probe descriptor we check if the euclidean distance from its nearest neighbor in the gallery is smaller than some threshold, and if so we classify it as the target photographer. We used Camera D1 sequences for training and D2 sequences for test. 16 randomly selected photographers were used for training the CNN, and the rest for verification. The same procedure was carried out for LPC (without training a CNN). Multiple 4s sequence predictions are aggregated using simple voting. The average EER for 12s sequences was 15.5\% (CNN) and 22\%(LPC). Although less accurate than trained classifiers, this shows the network learns to identity features that are general and can be transferred to identify unseen photographers. Nearest Neighbors classification on the optical flow raw features yielded very low performance in accordance with the findings of \cite{poleghead,yonetani2015egosurf}.  

\textit{Verification on FPSI:} We tried learning verification classifiers by choosing one photographer from the FPSI dataset as target, and 4 other photographers as negative training data. The morning sequences of the target photographer were used for training and the afternoon for testing. We tested the classification performance between the afternoon sequences of the target photographer and the remaining 6th non-target photographer from the FPSI dataset. The network however, fit to the train non-target photographers and has not been able to generalize to the unseen probe photographer. We therefore conclude that a significant number of photographers (such as present in the EVPR dataset) is required for training a verification classifier.

\begin{figure}[tb]
\centering{
\includegraphics[width=0.15\textwidth]{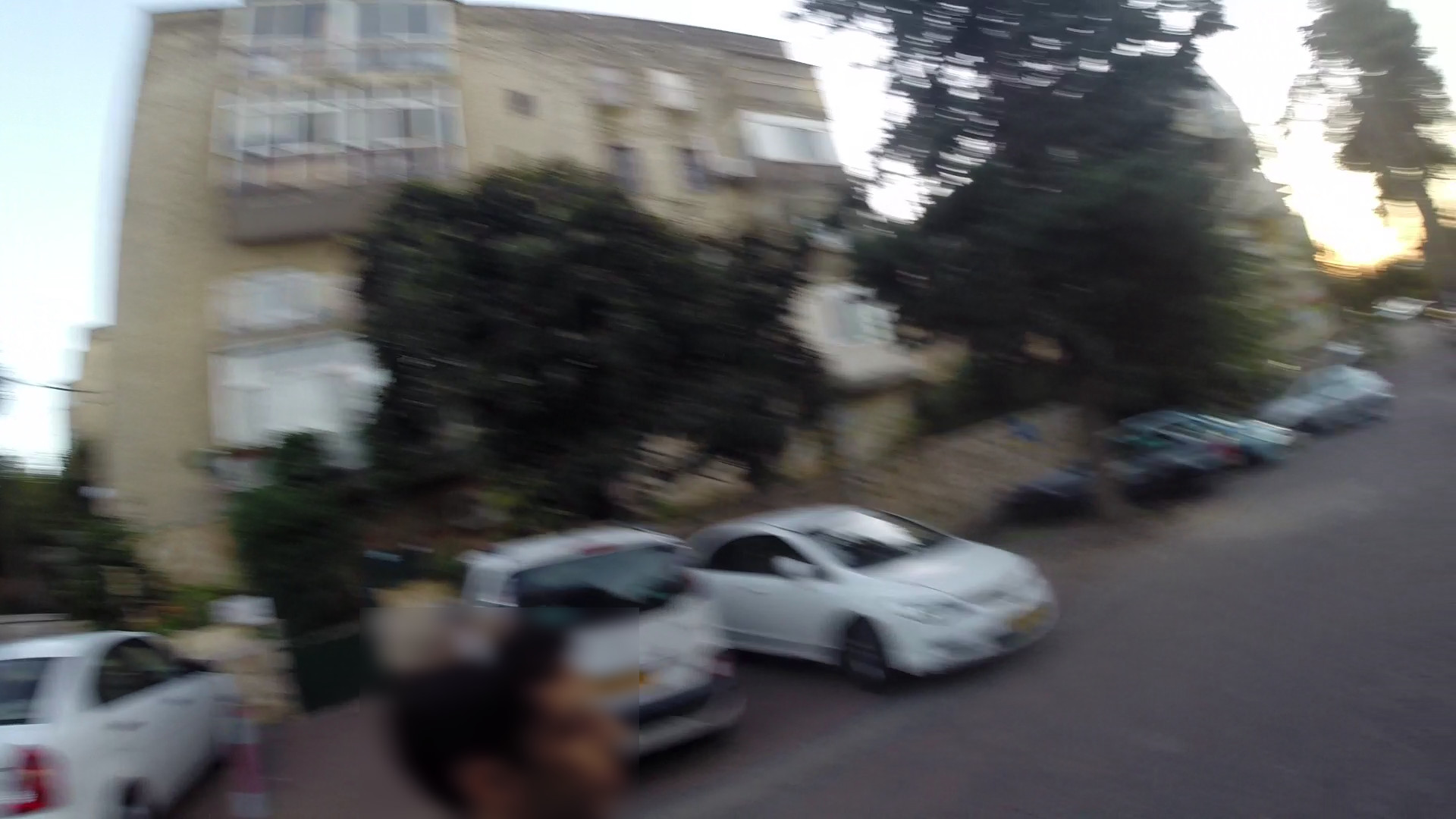}
\includegraphics[width=0.15\textwidth]{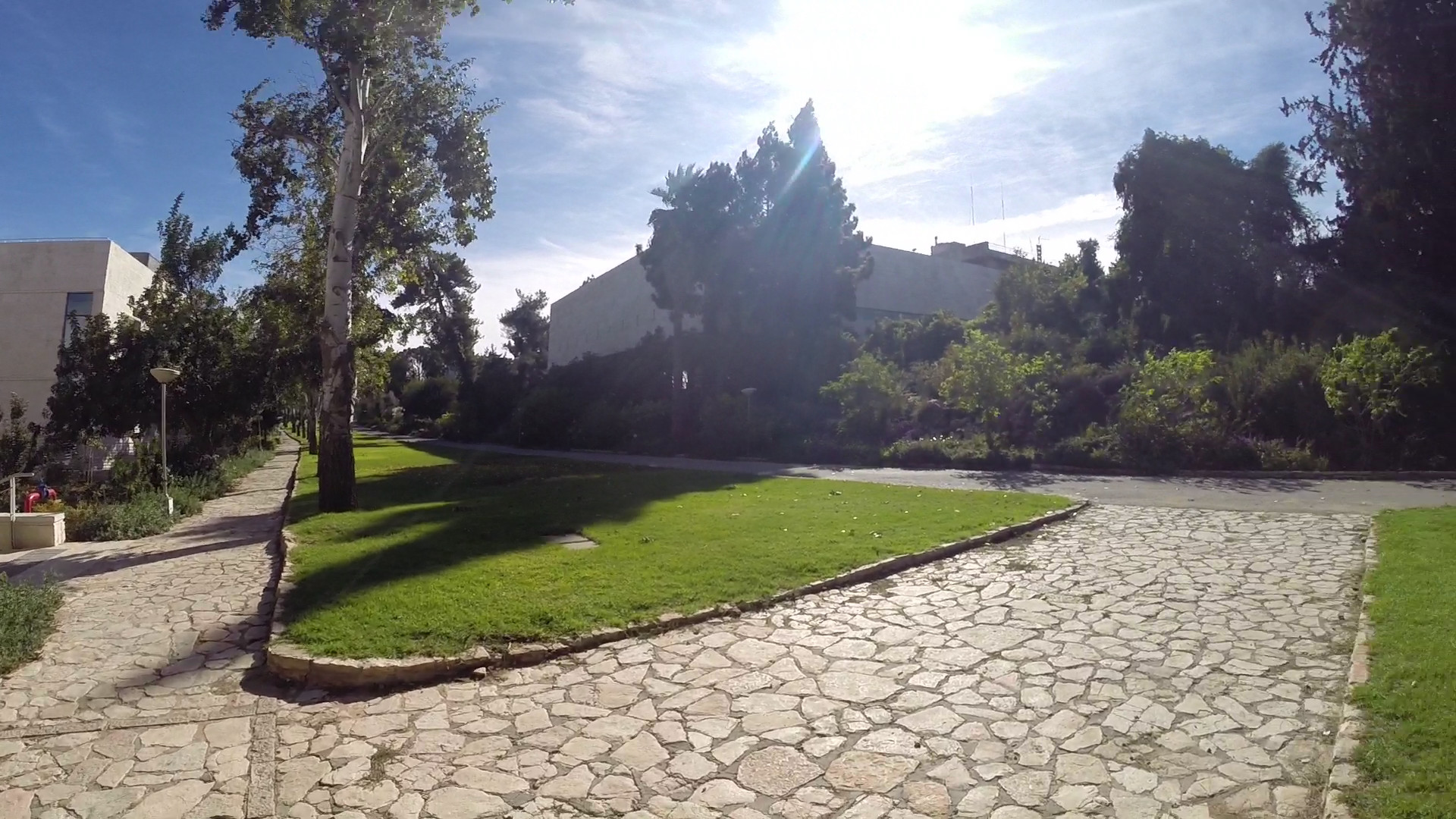}
\includegraphics[width=0.15\textwidth]{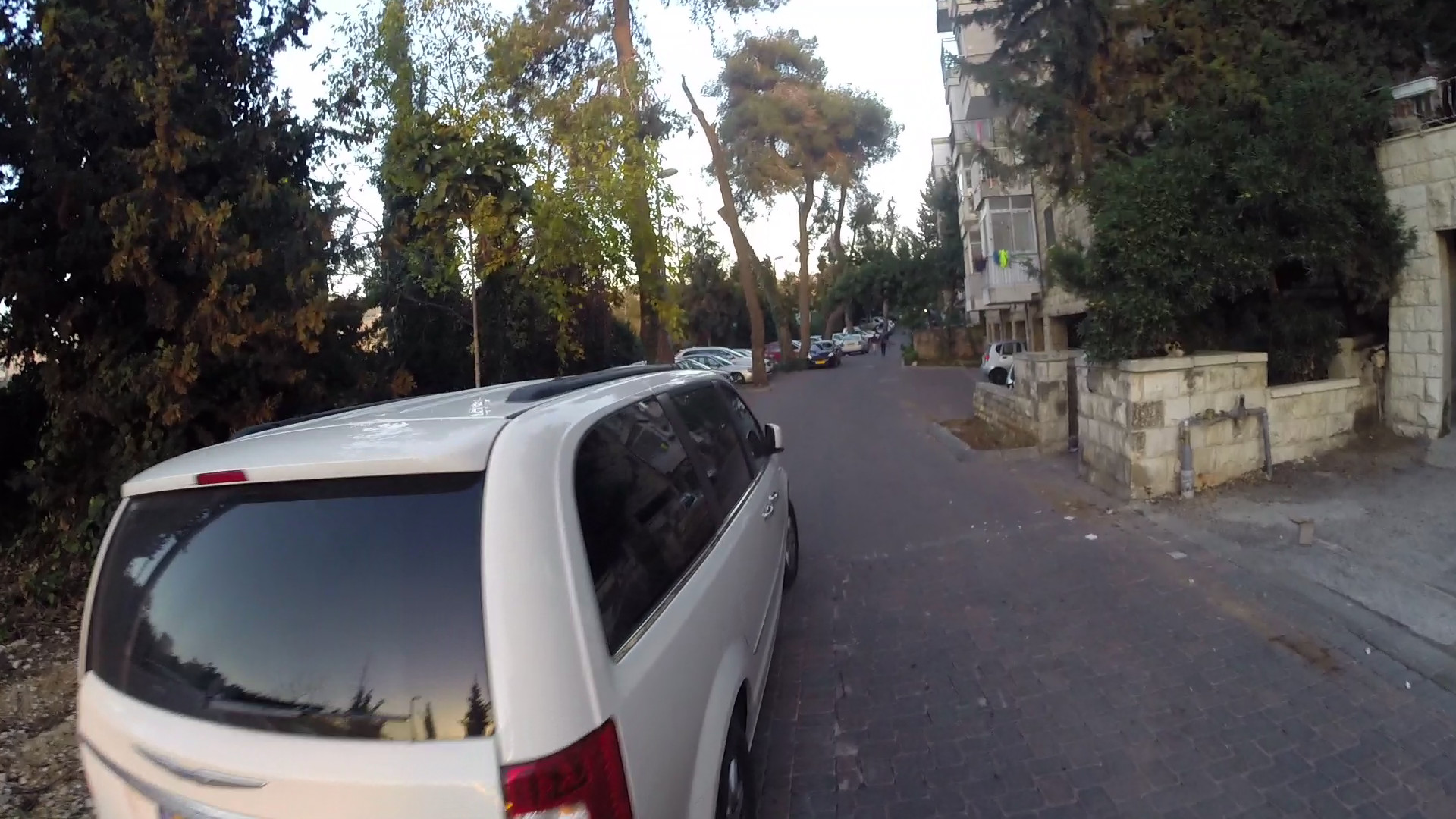}
a)~~~~~~~~~~~~~~~~~~~~~~~~~~~~~~~~~~b)~~~~~~~~~~~~~~~~~~~~~~~~~~~~~~~~~~c)
\caption{Common failure cases for the 4-second descriptor: a-b) Sharp turns of the head result in atypical fast motions, sometimes causing motion blur. c) Large moving objects can also cause atypical optical flow patterns.}
\label{Fig:Fail}}
\end{figure} 

\textit{Failure cases:} In Fig.~\ref{Fig:Fail} several cases are shown where the 4 second descriptor failed to give correct recognition. Failure can be caused by sharp head movements (sometimes causing significant blur), by large moving objects, or by lack of features for optical flow computation. It is likely that by identifying such cases and removing their descriptors, higher recognition performance may be achieved.

\section{Experimental Procedure}
\label{sec:pro}

In this section we give a detailed description of the experimental procedure used in Sec.~\ref{sec:results}.

\subsection{Dataset Description}
\label{sec:dsdes}

Two datasets were used for evaluation: a public general purpose dataset (FPSI) and a larger dataset (EVPR) collected by us to overcome some of the weaknesses of FPSI.

\subsubsection{FPSI Dataset}
\label{subsec:fpsi}

The First-Person Social Interactions (FPSI) dataset was collected by Fathi et al. \cite{fathi2012social} for the purpose of activity analysis. 6 individuals (5 males, 1 female) recorded a day's worth of egocentric video each using head-worn GoPro cameras. Due to battery and memory limitations of the camera, the photographers occasionally took the cameras off and put them on again, ensuring that camera extrinsic parameters were not kept constant. 

In this work we learn to recognize video photographers while walking, rather than sitting or standing. We therefore extracted the walking portions of each video using manual labels. It is possible to use a classifier such as described in \cite{polegtemporal} to find the walking intervals.

\subsubsection{EVPR Dataset}
\label{subsec:evb}

The FPSI dataset suffers from several drawbacks: it contains video only for a small number of photographers (6) and each photographer wears the same hat and camera all the time. It is therefore conceivable that learning camera parameters can help recognition. To overcome these issues we collected a larger dataset - Egocentric Video Photographer Recognition (EVPR). 

\begin{figure}[tb]
\centering{
\includegraphics[height=0.15\textwidth,width=0.1\textwidth]{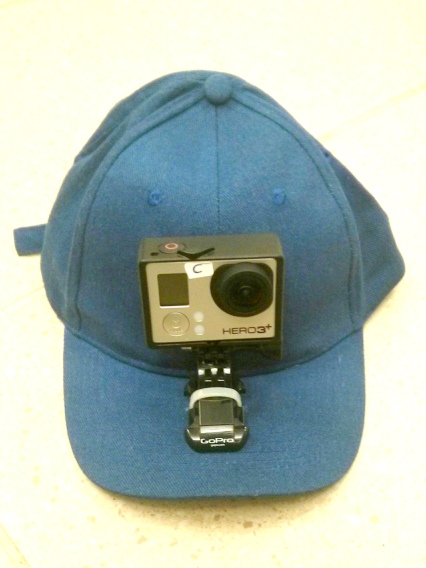}

\caption{The apparatus used to record the EVPR dataset.}
\label{Fig:Hat}}
\end{figure}

The EVPR consists of head-mounted video sequences collected from 32 photographers. Each video sequence was recorded with a GoPro camera attached to a baseball cap worn on the photographer's head (as in Fig.~\ref{Fig:Hat}). Each photographer was asked to walk normally for around 7 minutes along the same road. All photographers recorded two 7 minute video sequences on a single day using two different cameras (and caps). 20 photographers also recorded another sequence a week later. The use of different cameras for different sequences came to ensure that motion rather than camera calibration is learned. No effort was made to ensure that the same shoes would be used on both days (and in fact several persons had changed shoes between sequences).

\subsection{Evaluation Protocol}
\label{sec:proto}

\subsubsection{Photographer Identification}
\label{subsec:per_class}

Photographer identification sets to recognize a photographer from a closed set of $M$ candidates. For this task it is assumed that we have training data from all subjects. 

We tested our method both on the FPSI the EVPR datasets. In the FPSI dataset we used for each individual the first 80\% of sequences (taken in the morning) for training, and the last 20\% sequences recorded in the afternoon for testing. This is done to reduce overfitting to a particular time or camera setup. Data were randomly sub-sampled to ensure equal number of examples for each photographer in both training and testing sets. The results are described in Sec.~\ref{sec:results}.

For the EVPR dataset we used sequences from Camera D1 for training. For testing we use both sequences from Camera D2 (taken on the same day) and Camera D3 (taken a week later, when available). The results on each camera are compared to analyze whether recognition performance degrades within a week.    

\subsubsection{Photographer Verification}
\label{subsec:rich_per_ver}

Given a target photographer with a few minutes of training data, and negative training examples by other non-target photographers, we verify whether a probe test video sequence was recorded by the target photographer. Recognition on longer sequences is done by combining the predictions from subsequent short sequences. 
As the FPSI dataset contains only 6 photographers it was not suitable for this task (this was elaborated upon in Sec.~\ref{sec:dis}) therefore only the EVPR dataset was used for evaluating performance on this task. 
For each of 32 photographers: i) photographer is designated target ii) we selected sequences of the target photographer and 15 non-target photographers (randomly selected) for training a binary classifier. All training sequences were 7 minutes (200 descriptors) long and were recorded by camera D1. iii) Another sequence recorded by the target photographer and the remaining 16 participants that were not used for training, were used to test the classifier. Test sequences were recorded by camera D2. iv) The ROC curve and EER ware computed. Average EER and ROC for all photographers is finally obtained. 
As each sequence contained about 200 descriptors this formed a significant test set. Care was taken to ensure that all photographers (apart from the target) would appear in the training or test datasets but not in both. This was done to ensure we did not overfit to specific non-target photographers. We replicated positive training examples to ensure equal numbers of negative and positive training and test data.

\subsection{Implementation Details}
\label{sec:imp}

\textit{Features:} In all experiments the optical flow grid size used was 10$\times$5. In the CNN experiments, all optical flow values were divided by the square-root of their absolute value, this was found to help performance by decreasing the significance of extreme values. Feature vectors of length 60 frames at 15 fps (4s) were used. Feature vectors were extracted every 2s (with a 2s overlap).

\textit{Normalization:} We followed the standard practice - For the LPC descriptor, all feature vectors were mean and variance normalized across the training set before being used by the SVM. For the CNN, feature vectors were mean subtracted before being input to the CNN.

\textit{Training:} The SVM was trained using LIBSVM \cite{CC01a}. We used $\sigma = 1e-4$ and $ C = 1$ for LPC, $C = 10$ for the raw features. The CNN was trained by AdaGrad \cite{duchi2011adaptive} with learning rate 0.01 on a GPU using the Caffe \cite{jia2014caffe} package. The mini-batch size was 200.

\section{Conclusion}
\label{sec:conc}
A method to recognize the photographer of head-worn egocentric camera video has been presented. We show that photographer identity can be found from body motion information as expressed in camera motion when walking. Recognition was done with both physically motivated hand designed descriptors, and with a Convolutional Neural Network. Both methods gave good recognition performance. The CNN classifier was shown to generalize and improve on the LPC hand-designed descriptor. 

The time-invariant CNN architecture presented here is quite general and can be used for other video classification tasks relying on coarse optical flow.


We have tested the effects of simple 2D video stabilization on classification accuracy, and found only slight degradation in performance. It is possible that more elaborate stabilization would have a greater effect.

The implication of our work is that photographers' head-worn egocentric videos give much information away. This information can be used benevolently (e.g. camera theft prevention, user analytics on video sharing websites) or maliciously. Care should therefore be taken when sharing such video.

\ifcvprfinal{
\pagestyle{empty}
~\\
\noindent
{\bf Acknowledgment.} This research was supported by Intel-ICRC and by the Israel Science Foundation.
}

{
\bibliographystyle{ieee}
\bibliography{egbib}

\begin{thebibliography}{10}\itemsep=-1pt

\bibitem{bar2014classification}
Y.~Bar, N.~Levy, and L.~Wolf.
\newblock Classification of artistic styles using binarized features derived
  from a deep neural network.
\newblock In {\em ECCV 2014 Workshops}, pages 71--84, 2014.

\bibitem{bengio2009learning}
Y.~Bengio.
\newblock Learning deep architectures for {AI}.
\newblock {\em Foundations and trends in Machine Learning}, 2(1):1--127, 2009.

\bibitem{campbell1997speaker}
J.~P. Campbell~Jr.
\newblock Speaker recognition: a tutorial.
\newblock {\em Proceedings of the IEEE}, 85(9):1437--1462, 1997.

\bibitem{CC01a}
C.-C. Chang and C.-J. Lin.
\newblock {LIBSVM}: A library for support vector machines.
\newblock {\em ACM Trans. on Intelligent Systems and Technology}, 2011.

\bibitem{duchi2011adaptive}
J.~Duchi, E.~Hazan, and Y.~Singer.
\newblock Adaptive subgradient methods for online learning and stochastic
  optimization.
\newblock {\em JMLR}, 2011.

\bibitem{fathi2012social}
A.~Fathi, J.~K. Hodgins, and J.~M. Rehg.
\newblock Social interactions: A first-person perspective.
\newblock In {\em CVPR}, 2012.

\bibitem{furui1981cepstral}
S.~Furui.
\newblock Cepstral analysis technique for automatic speaker verification.
\newblock {\em ICASSP}, 1981.

\bibitem{hays2008im2gps}
J.~Hays, A.~Efros, et~al.
\newblock Im2gps: estimating geographic information from a single image.
\newblock In {\em CVPR'08}, pages 1--8, 2008.

\bibitem{jia2014caffe}
Y.~Jia, E.~Shelhamer, J.~Donahue, S.~Karayev, J.~Long, R.~Girshick,
  S.~Guadarrama, and T.~Darrell.
\newblock Caffe: Convolutional architecture for fast feature embedding.
\newblock {\em arXiv:1408.5093}, 2014.

\bibitem{johnson2008image}
C.~R. Johnson~Jr, E.~Hendriks, I.~J. Berezhnoy, E.~Brevdo, S.~M. Hughes,
  I.~Daubechies, J.~Li, E.~Postma, and J.~Z. Wang.
\newblock Image processing for artist identification.
\newblock {\em IEEE Signal Processing Magazine}, 25(4):37--48, 2008.

\bibitem{kitani2011fast}
K.~M. Kitani, T.~Okabe, Y.~Sato, and A.~Sugimoto.
\newblock Fast unsupervised ego-action learning for first-person sports videos.
\newblock In {\em CVPR}, 2011.

\bibitem{krizhevsky2012imagenet}
A.~Krizhevsky, I.~Sutskever, and G.~E. Hinton.
\newblock Imagenet classification with deep convolutional neural networks.
\newblock In {\em NIPS}, 2012.

\bibitem{lee2015predicting}
S.~Lee, H.~Zhang, and D.~J. Crandall.
\newblock Predicting geo-informative attributes in large-scale image
  collections using convolutional neural networks.
\newblock In {\em WACV'15}, pages 550--557, 2015.

\bibitem{lu2013story}
Z.~Lu and K.~Grauman.
\newblock Story-driven summarization for egocentric video.
\newblock In {\em CVPR}, 2013.

\bibitem{lucas1981iterative}
B.~D. Lucas, T.~Kanade, et~al.
\newblock An iterative image registration technique with an application to
  stereo vision.
\newblock In {\em IJCAI}, volume~81, pages 674--679, 1981.

\bibitem{mantyjarvi2005identifying}
J.~Mantyjarvi, M.~Lindholm, E.~Vildjiounaite, S.-M. Makela, and H.~Ailisto.
\newblock Identifying users of portable devices from gait pattern with
  accelerometers.
\newblock In {\em ICASSP}, 2005.

\bibitem{murase1996moving}
H.~Murase and R.~Sakai.
\newblock Moving object recognition in eigenspace representation: gait analysis
  and lip reading.
\newblock {\em Pattern Recognotion Letters}, 17(2):155--162, 1996.

\bibitem{nishino2006corneal}
K.~Nishino and S.~K. Nayar.
\newblock Corneal imaging system: Environment from eyes.
\newblock {\em IJCV}, 70(1):23--40, 2006.

\bibitem{polegtemporal}
Y.~Poleg, C.~Arora, and S.~Peleg.
\newblock Temporal segmentation of egocentric videos.
\newblock In {\em CVPR}.

\bibitem{poleghead}
Y.~Poleg, C.~Arora, and S.~Peleg.
\newblock Head motion signatures from egocentric videos.
\newblock In {\em ACCV}, 2014.

\bibitem{reynolds1995robust}
D.~A. Reynolds and R.~C. Rose.
\newblock Robust text-independent speaker identification using gaussian mixture
  speaker models.
\newblock {\em IEEE Trans. on Speech and Audio Processing}, 3(1):72--83, 1995.

\bibitem{ryoo2013first}
M.~S. Ryoo and L.~Matthies.
\newblock First-person activity recognition: What are they doing to me?
\newblock In {\em CVPR}, 2013.

\bibitem{shiraga2012gait}
K.~Shiraga, N.~T. Trung, I.~Mitsugami, Y.~Mukaigawa, and Y.~Yagi.
\newblock Gait-based person authentication by wearable cameras.
\newblock In {\em INSS}, 2012.

\bibitem{thomas2015s}
C.~Thomas and A.~Kovashka.
\newblock Who's behind the camera? identifying the authorship of a photograph.
\newblock {\em arXiv:1508.05038}, 2015.

\bibitem{yonetani2015egosurf}
R.~Yonetani, K.~M. Kitani, and Y.~Sato.
\newblock Ego-surfing first person videos.
\newblock In {\em CVPR}, 2015.

\end{thebibliography}
}

\end{document}